\documentclass{article}

\usepackage{arxiv}

\usepackage[utf8]{inputenc} 
\usepackage[T1]{fontenc}    
\usepackage{hyperref}       
\usepackage{url}            
\usepackage{booktabs}       
\usepackage{amsfonts}       
\usepackage{nicefrac}       
\usepackage{microtype}      
\usepackage{lipsum}
\usepackage{graphicx}
\graphicspath{ {./images/} }
\usepackage{subfigure}

\usepackage{algorithm}
\usepackage{algpseudocode}

\title{Temperature Matters: Enhancing Watermark Robustness Against Paraphrasing Attacks}
\date{}

\author{
 Badr Youbi Idrissi\thanks{All authors contributed equally to this work.}\\
  \texttt{badryoubiidrissi@gmail.com} \\
   \And
 Monica Millunzi\footnotemark[1]\\
  \texttt{monica.millunzi@unimore.it} \\
  \And
 Amelia Sorrenti\footnotemark[1]\\
  \texttt{amelia.sorrenti@phd.unict.it} \\
    \And
 Lorenzo Baraldi\footnotemark[1]\\
  \texttt{lorenzo.baraldi@phd.unipi.it} \\
    \And
 Daryna Dementieva\footnotemark[1]\\
  \texttt{daryna.dementieva@tum.de} \\
}
\vspace{0.2cm}

\begin{document}
\maketitle
\begin{abstract}
In the present-day scenario, Large Language Models (LLMs) are establishing their presence as powerful instruments permeating various sectors of society. While their utility offers valuable support to individuals, there are multiple concerns over potential misuse. Consequently, some academic endeavors have sought to introduce watermarking techniques, characterized by the inclusion of markers within machine-generated text, to facilitate algorithmic identification.
This research project is focused on the development of a novel methodology for the detection of synthetic text, with the overarching goal of ensuring the ethical application of LLMs in AI-driven text generation. The investigation commences with replicating findings from a previous baseline study~\cite{aarson}, thereby underscoring its susceptibility to variations in the underlying generation model. Subsequently, we propose an innovative watermarking approach and subject it to rigorous evaluation, employing paraphrased generated text to asses its robustness. Experimental results highlight the robustness of our proposal compared to the~\cite{aarson} watermarking method.

\end{abstract}


\vspace{0.3cm}
\section{Introduction}
\label{sec:intro}
In an era dominated by Large Language Models (LLMs), we stand at the intersection of unprecedented linguistic capabilities and profound ethical challenges. While the potential applications of LLMs are vast and promising, their exponential growth also brings to the forefront the serious concerns associated with their potential misuse \cite{bergman2022guiding,mirsky2022threat,kertysova2018artificial}.
From election rigging to social engineering campaigns using automated bots on social media platforms, from spreading fake news to dishonest use in academic and coding assignments, the dark side of LLMs has become increasingly apparent. 
As a result, the line between genuine human-generated content and machine-generated text is blurring, making accountability and transparency in the area of LLM-generated text urgent ~\cite{bender2021dangers,crothers2023machine,grinbaum2022ethical}.

To address this critical issue, our project explores the concept of watermarking, which involves the subtle incorporation of imperceptible markers, known as \emph{watermarks}, into machine-generated text. Watermarking enables the algorithmic identification of machine-generated text while remaining largely undetectable to human readers \cite{kirchenbauer2023reliability,kirchenbauer2023watermark,wen2023tree}. The aim is not only to facilitate the identification of machine-generated text, but also to minimize false positives, ensuring that authentic human content remains unaltered.



To summarize, the main contributions of our work are the following:
\begin{itemize}
    \item We replicate the watermarking detection results presented in \cite{aarson} with open-source model.
    \item We propose a new approach for watermarking which changes the temperature $T$ of sampling at each token.
    \item We test our approach on the robustness to the paraphrasing attack, i.e. if some part of the watermarked text is paraphrased.
\end{itemize}

\begin{figure}[htbp]
    \centering
    \begin{minipage}[t]{0.48\textwidth}
        \centering
        \includegraphics[width=\textwidth]{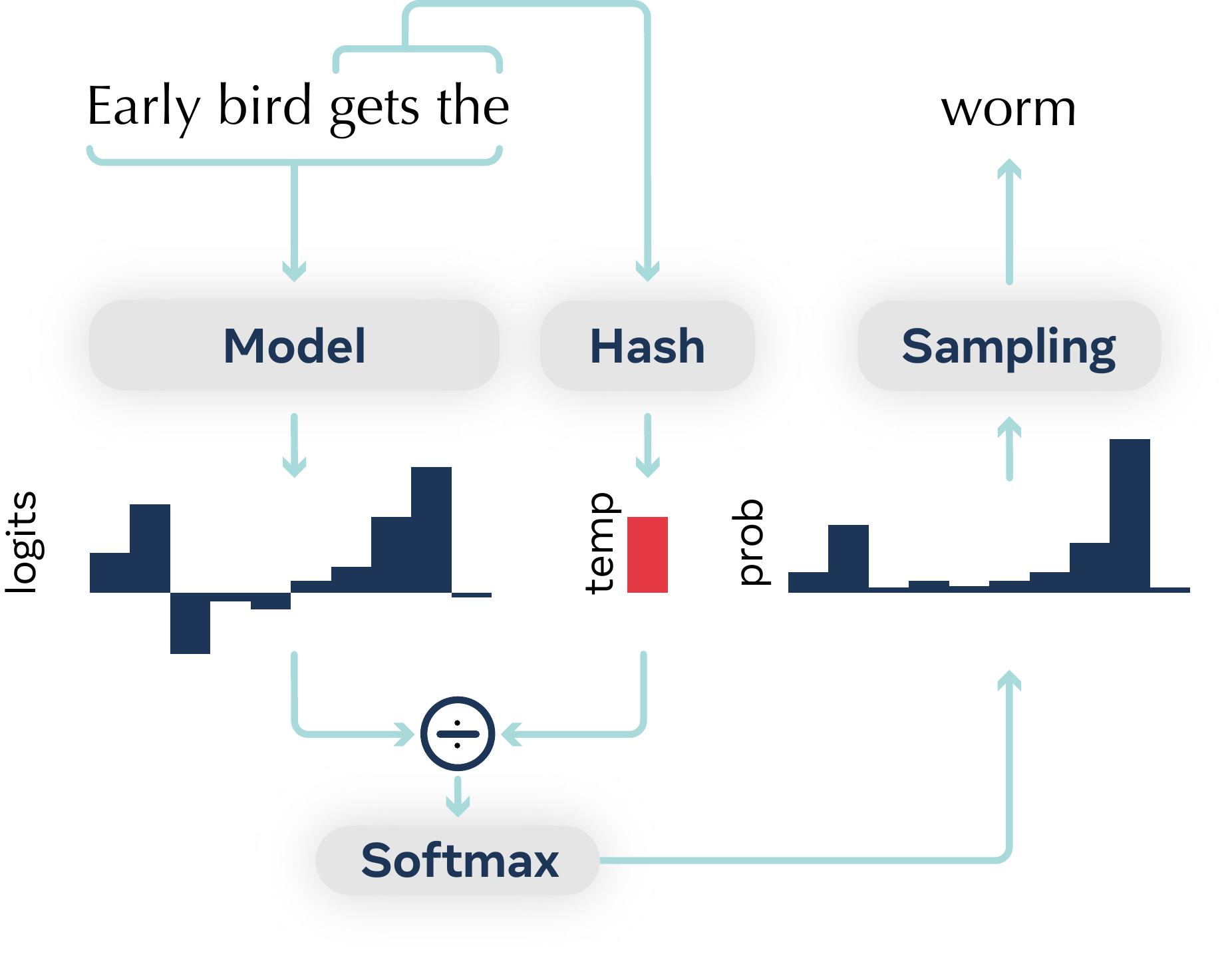}
        \caption{Overview of our temperature-based watermarking technique: we employ $h$ tokens (in this example, $2$) as the seed for a hashing process to sample a temperature parameter. This temperature parameter is subsequently applied to adjust the logits, for the prediction of the subsequent token.}
        \label{fig:bert_paraphrasing}
    \end{minipage}
    \hfill
    \begin{minipage}[t]{0.40\textwidth}
        \centering
        \includegraphics[width=\textwidth]{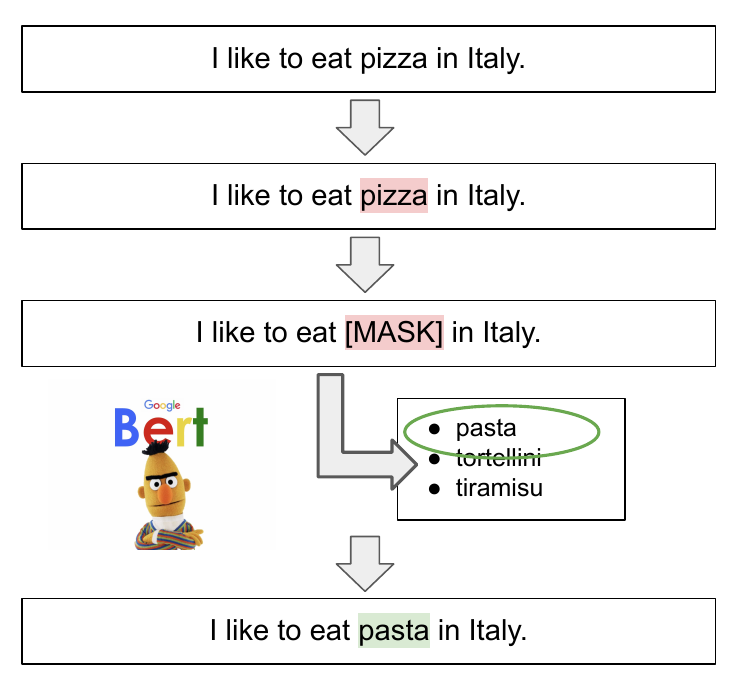}
        \caption{An example of text paraphrasing attack utilizing BERT~\cite{alaparthi2020bidirectional}, i.e. substituting randomly selected words with model predictions.}
        \label{fig:bert_paraphrasing}
    \end{minipage}
\end{figure}

\section{Method}
\label{sec:method}

In this section, we present our novel watermarking technique tailored for Large Language Models. To provide a comprehensive delineation of our approach, we commence by establishing the necessary notational framework and subsequently elaborate on the employed architectural configurations.

\subsection{Notation and Preliminaries}

Given a word vocabulary of $|V|$ granularity, the Large Language Model (LLM) for next-word prediction is a function $f_{\theta}$, often parameterized by a neural network, which takes in input a sequence 
T of tokens denoted as ${s^{(t)}}\in V^T$ and outputs a vector of $V$ logits, one for each word in the vocabulary. These logits are then passed through a softmax operator to convert them into a discrete probability distribution over the vocabulary. The next token is commonly sampled from this distribution using a sampling strategy of choice. For notational purposes, tokens with negative indices, $s^{(-N_p)},\cdots, s^{(-1)}$, indicate a prompt characterized by a length of $N_p$, while $s^{(0)},\cdots,s^{(T)}$ correspond to tokens produced by an AI system as a response to the given prompt.

\subsection{Model}
In this project, we employed the Vicuna-7B model~\cite{chiang2023vicuna}, which is a LLaMA~\cite{touvron2023llama} model fine-tuned on user-shared conversations collected from ShareGPT, a ChatGPT dialogue corpus crawled
from sharegpt.com.

\subsection{Watermarking Generation}
While~\cite{kirchenbauer2023watermark} watermarking method randomly partitioned the vocabulary into two lists of equal size, namely the ``green list" and the ``red list", we have opted for an alternative methodology by exclusively considering the logits vector obtained as the model's output. 

Our approach uses a context window of size $h$ to generate a unique hash. This hash is then used as a seed for generating a random temperature $T^{t}$ for each token. This temperature is sampled as follows :

\begin{equation}
T^{t} \sim T_0(m+(M-m)\mathcal{U}^{t})
\end{equation}

Where $U$ is a uniform pseudo random variable seeded by the hash of the tokens $[s^{t-h},\dots,s^{t-1}]$. In other words, U is a deterministic function of the context window. This will allow us to recompute the same temperatures when detecting the watermark later on. The temperature varies around the base value $T_0$. For some tokens the temperature will be higher and for others it will be lower. We can then compute the probabilities of each token in the vocabulary using the softmax function.
\begin{equation}
P(s^{t}=k|s^{1},\dots,s^{t-1}; T^{t}) = \frac{\exp(l^{(t)}_k/T^{t})}{\sum_i{\exp(l^{(t)}_i/T^{t}})}
\end{equation}

Since the temperature controls the entropy of the output distribution, this forces the model to take a specific path during generation that allows us to distinguish it from human generated text later on.

The resulting vector of probabilities is then used for sampling with any desired method. A detailed version of the algorithm is presented in Alg.~\ref{hard}.

\begin{algorithm}[t]
   \caption{Proposed Temperature Watermarking}
   \label{hard}
\begin{algorithmic}
\State \textbf{Input:} prompt, $s^{(-N_p)},\cdots, s^{(-1)}$,\\n.o. tokens to use for hashing $h$, model $f_{\theta}$ 
\For{$t=0,1,\cdots, t$}
\begin{enumerate}
\item Apply the language model to prior tokens $s^{(-N_p)}\cdots s^{(t-1)}$ and retain the logits $l^{(t)}$ over the vocabulary.
\item Compute a hash of the last $h$ tokens of the prompt $s^{(t-h)}\cdots\ s^{(t-1)}$ and use it to seed a random number generator.
\item Using this seed, randomly generate a temperature value $T^{t} \sim T_0(m+(M-m)\mathcal{U}^{t})$
\item Rescale the logits $l^{(t)} = l^{(t)}/{T^{t}}$
\item Convert the logits to a probability vector using the softmax operator $p^{(t)}_k = \exp(l^{(t)}_k)/\sum_i \exp(l^{(t)}_i).$
\item Sample using this probability distribution

\end{enumerate}
\EndFor
\end{algorithmic}
\end{algorithm}

\subsection{Watermarking Detection}
The outcome of the process described above is that the model's confidence in its predictions varies in a known way throughout generation. This variation in confidence across a specific sequence of tokens is what ultimately constitutes the watermark.

To detect the watermark, it is sufficient to recompute the same probabilities as in the watermarking phase, by doing a forward pass on the text, then select the probabilities of the tokens, and average them. Eq.~\ref{eq:out}.

\begin{equation}
\label{eq:out}
    \mathrm{Score} = \frac{1}{N}\sum_{t=1}^{N} P(s^{t}|s^{1},\dots,s^{t-1}; T^{t})
\end{equation}
This score can be then used to determine whether the sequence is watermarked or not. Indeed, sequences with no watermarking are likely to have a smaller score than watermarked sequences.



\begin{figure*}[t]
    \centering
    \includegraphics[width=0.99\textwidth]{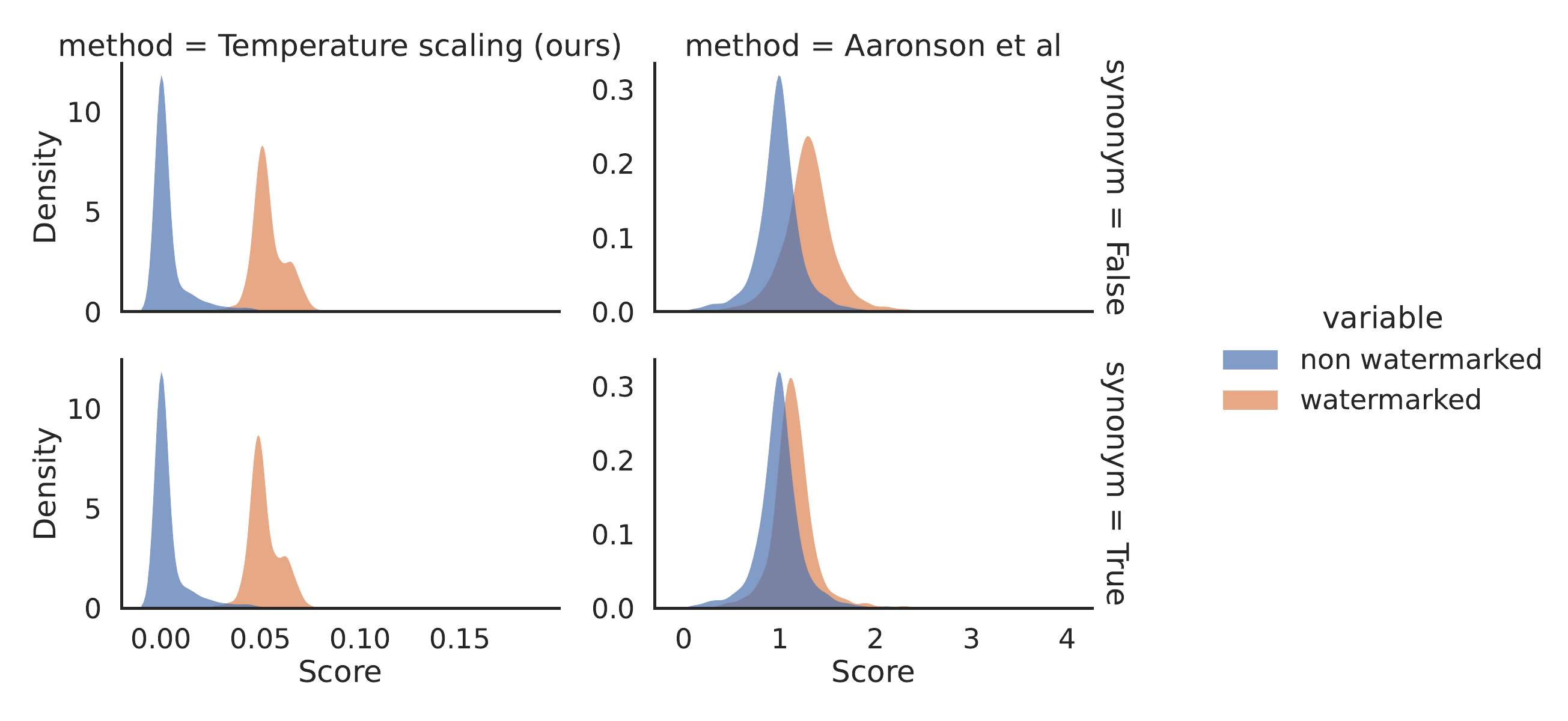}
    \caption{Score of our temperature-based watermarking (left column) technique compared to the baseline method (right column). In the first row both methods are compared without paraphrasing attacks, on the second row BERT~\cite{alaparthi2020bidirectional} is applied to 30\% of the generated input tokens.}
    \label{fig:results_with_attacks}
\end{figure*}

\section{Evaluation Setup}
\subsection{Baseline Reproducibility}
Firstly, we decide to reprocude the machine-generated papers detection with already published SOTA approach. For this, as for baseline, we refer to the method presented in \cite{aarson}. There, the experiments were conducted on LLaMa~\cite{touvron2023llama} which is available upon the request. We substitute the model with the opensource model---Vicuna\footnote{\href{https://huggingface.co/lmsys/vicuna-7b-v1.3}{https://huggingface.co/lmsys/vicuna-7b-v1.3}}---with has the same amount of hyper-parameters and fairly compared performance on the benchmarks.

Thus, with this experiment, we want to test the following \textit{\textbf{Hypothesis H1}}: if the published results of machine-generated texts detection are reproducible with the open-source model. While in \cite{kirchenbauer2023watermark} the theoretical proof of watermarking model-agnosticism was provided, it still should be verified empirically.

\subsection{Paraphrasing Attack}

One of the main attacks on the watermarking can be the paraphrasing of some parts of machine-generated watermarked text \cite{kirchenbauer2023reliability}. In \cite{DBLP:journals/corr/abs-2308-00113}, the significant drop of their detection method performance on the paraphrased texts was observed. Thus, we formulate the next \textit{\textbf{Hypothesis H2}}: if our proposed watermarking method is robust to the paraphrasing attacks.

To imitate the paraphrasing process, we utilize BERT\footnote{\href{https://huggingface.co/bert-base-uncased}{https://huggingface.co/bert-base-uncased}}  \cite{alaparthi2020bidirectional} for masked language modeling (Figure~\ref{fig:bert_paraphrasing}). For each text sample, we replace 30\% of the tokens with \texttt{[MASK]} and ask the model for the prediction. Each \texttt{[MASK]} prediction was done one-by-one to save the content.

\subsection{Dataset and Metric}

For the test set, we use random 1k samples fro Alpaca dataset.\footnote{\href{https://huggingface.co/datasets/tatsu-lab/alpaca}{https://huggingface.co/datasets/tatsu-lab/alpaca}} This dataset consists of the prompts, human answers, and machine-generated texts. We generate our own texts with Vicuna and the proposed watermarking method inputting the prompts and comparing them with human-written ones from the dataset. 

We report the F1 score, True Positive Rate (TPR), and False Positive Rate (FPR) as a main evaluation metric as with our work we are aiming to increase the probability of machine-generated texts detection in comparison to the previous approach.

\begin{figure*}[th!]
  \centering
    \centering
    \includegraphics[width=0.85\linewidth]{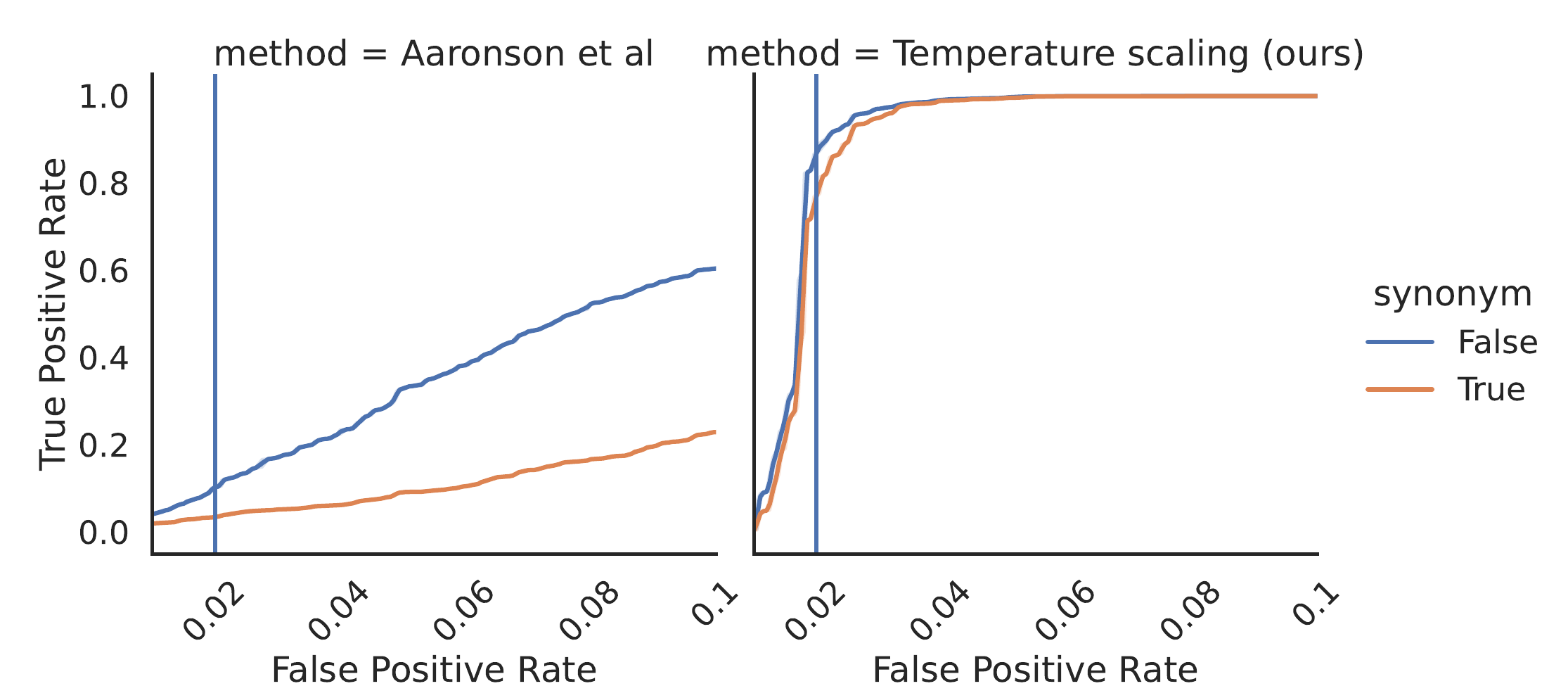}
    \label{fig:baseline_comparison_a}
  \hfill
  \caption{Receiver Operating Characteristic (ROC) curve for the baseline~\cite{aarson} approach (left) and our proposed approach (right). \textit{synonym} indicate if the paraphrasing attack was applied or not.}
 \label{fig:baseline_comparison_roc}
\end{figure*}

\section{Results}
\label{sec:results}

Firstly, we illustrate the comparison with the baseline approach on the first row of Figure~\ref{fig:results_with_attacks}. We can observe that in the baseline approach (right column) it is difficult to differentiate between distributions for human-written and machine-generated text. On the other hand in the proposed approach (left column), the differences in the distributions are clear making it easy to separate machine-generated textual descriptions. Thus, previously stated \textit{\textbf{H1}} is not confirmed. The reasons for this are the grounds for further investigation.

For \textbf{H2}, we are investigating the robustness to the paraphrasing attack of the proposed technique. The intuition in the proposed watermark technique lies in the idea that the change in temperature will, at some point, allow the model to take into account not only the most probable word on the step but also its synonyms. In the second row of Figure~\ref{fig:results_with_attacks}, the analysis with attacked generated texts is represented. In this case, the previous observation on the distinction of human and machine-generate texts is even more accentuated. Indeed our model seems invariant to the attack while baseline-processed text is nearly overlapped with human input making distinction difficult.

Furthermore, a quantitative assessment has been carried out using the Receiver Operating Characteristic (ROC) curve. As depicted in Figure~\ref{fig:baseline_comparison_roc}, it is evident that the temperature watermarking method we propose outperforms the baseline approach at all stages of evaluation. To illustrate, when the False Positive Rate (FPR) is held constant at 2\%, the baseline method achieves a True Positive Rate (TPR) of 15\%, whereas our proposed method attains a TPR of 90\%. Moreover, this performance advantage holds true even when subjected to a paraphrasing attack. This proves \textbf{H2}---our proposed temperature watermark is more robust to the paraphrasing attack---making further analysis possible in future works.

\section{Conclusion}
%
%

\label{sec:conclusion}
In this work, we investigated the robustness of the watermarking approach for machine-generated texts. Firstly, we found out that reproducibility of the detection results based on one model is challenging on the other model while the method is claimed to be model agnostic. 

Then, we proposed a new watermark which generates a unique temperature at each step of generation. This approach made the watermark detection easier and more robust on the Vicuna model. We also considered paraphrasing attack as one of the main attacks on the machine-generated texts detector. In our proposed watermarking method, this attack is addressed outperforming the baseline. The method allows to choose at some steps within more options thus take into account more suitable generation options. We understand that the thorough investigation of the proposed method is still to be addressed in the future work.


\begin{thebibliography}{10}

\bibitem{aarson}
S.~Aaronson and H.~Kirchner.
\newblock Watermarking gpt outputs.
\newblock \url{https://www.scottaaronson.com/talks/watermark.ppt}, 2023.

\bibitem{bergman2022guiding}
A~Stevie Bergman, Gavin Abercrombie, Shannon Spruit, Dirk Hovy, Emily Dinan, Y-Lan Boureau, Verena Rieser, et~al.
\newblock Guiding the release of safer e2e conversational ai through value sensitive design.
\newblock In {\em Proceedings of the 23rd Annual Meeting of the Special Interest Group on Discourse and Dialogue}. Association for Computational Linguistics, 2022.

\bibitem{mirsky2022threat}
Yisroel Mirsky, Ambra Demontis, Jaidip Kotak, Ram Shankar, Deng Gelei, Liu Yang, Xiangyu Zhang, Maura Pintor, Wenke Lee, Yuval Elovici, et~al.
\newblock The threat of offensive ai to organizations.
\newblock {\em Computers \& Security}, page 103006, 2022.

\bibitem{kertysova2018artificial}
Katarina Kertysova.
\newblock Artificial intelligence and disinformation: How ai changes the way disinformation is produced, disseminated, and can be countered.
\newblock {\em Security and Human Rights}, 29(1-4):55--81, 2018.

\bibitem{bender2021dangers}
Emily~M Bender, Timnit Gebru, Angelina McMillan-Major, and Shmargaret Shmitchell.
\newblock On the dangers of stochastic parrots: Can language models be too big?
\newblock In {\em Proceedings of the 2021 ACM conference on fairness, accountability, and transparency}, pages 610--623, 2021.

\bibitem{crothers2023machine}
Evan Crothers, Nathalie Japkowicz, and Herna~L Viktor.
\newblock Machine-generated text: A comprehensive survey of threat models and detection methods.
\newblock {\em IEEE Access}, 2023.

\bibitem{grinbaum2022ethical}
Alexei Grinbaum and Laurynas Adomaitis.
\newblock The ethical need for watermarks in machine-generated language.
\newblock {\em arXiv preprint arXiv:2209.03118}, 2022.

\bibitem{kirchenbauer2023reliability}
John Kirchenbauer, Jonas Geiping, Yuxin Wen, Manli Shu, Khalid Saifullah, Kezhi Kong, Kasun Fernando, Aniruddha Saha, Micah Goldblum, and Tom Goldstein.
\newblock On the reliability of watermarks for large language models.
\newblock {\em arXiv preprint arXiv:2306.04634}, 2023.

\bibitem{kirchenbauer2023watermark}
John Kirchenbauer, Jonas Geiping, Yuxin Wen, Jonathan Katz, Ian Miers, and Tom Goldstein.
\newblock A watermark for large language models.
\newblock {\em arXiv preprint arXiv:2301.10226}, 2023.

\bibitem{wen2023tree}
Yuxin Wen, John Kirchenbauer, Jonas Geiping, and Tom Goldstein.
\newblock Tree-ring watermarks: Fingerprints for diffusion images that are invisible and robust.
\newblock {\em arXiv preprint arXiv:2305.20030}, 2023.

\bibitem{alaparthi2020bidirectional}
Shivaji Alaparthi and Manit Mishra.
\newblock Bidirectional encoder representations from transformers (bert): A sentiment analysis odyssey.
\newblock {\em arXiv preprint arXiv:2007.01127}, 2020.

\bibitem{chiang2023vicuna}
Wei-Lin Chiang, Zhuohan Li, Zi~Lin, Ying Sheng, Zhanghao Wu, Hao Zhang, Lianmin Zheng, Siyuan Zhuang, Yonghao Zhuang, Joseph~E Gonzalez, et~al.
\newblock Vicuna: An open-source chatbot impressing gpt-4 with 90\%* chatgpt quality.
\newblock {\em See https://vicuna. lmsys. org (accessed 14 April 2023)}, 2023.

\bibitem{touvron2023llama}
Hugo Touvron, Thibaut Lavril, Gautier Izacard, Xavier Martinet, Marie-Anne Lachaux, Timothée Lacroix, Baptiste Rozière, Naman Goyal, Eric Hambro, Faisal Azhar, Aurelien Rodriguez, Armand Joulin, Edouard Grave, and Guillaume Lample.
\newblock Llama: Open and efficient foundation language models, 2023.

\bibitem{DBLP:journals/corr/abs-2308-00113}
Pierre Fernandez, Antoine Chaffin, Karim Tit, Vivien Chappelier, and Teddy Furon.
\newblock Three bricks to consolidate watermarks for large language models.
\newblock {\em CoRR}, abs/2308.00113, 2023.

\end{thebibliography}
\end{document}